\documentclass[10pt,twocolumn,letterpaper]{article}

\pdfoutput=1

\usepackage{iccv}
\usepackage{times}
\usepackage{epsfig}
\usepackage{graphicx}
\usepackage{amsmath}
\usepackage{amssymb}

\usepackage{caption} % added by fei
\usepackage{enumerate} % added by fei
\usepackage{verbatim} % added by fei 
\usepackage{multirow} % added by fei.
\usepackage{subfigure} % added by fei. for subfigures with captions
\usepackage{array} % added by fei
\usepackage{color} % added by fei
\newcommand\abs[1]{\left|#1\right|} % added by fei

\newcommand{\argmin}{\operatornamewithlimits{argmin}} % added by fei. write argmin/argmax properly:
\newcommand{\argmax}{\operatornamewithlimits{argmax}} % centre the variable under the whole word

\usepackage[pagebackref=true,breaklinks=true,letterpaper=true,colorlinks,bookmarks=false]{hyperref}

\iccvfinalcopy % *** Uncomment this line for the final submission

\def\iccvPaperID{1785} % *** Enter the ICCV Paper ID here

\ificcvfinal\fi
\begin{document}

%%%%%%%%% TITLE
\title{Person Re-Identification with Vision and Language}

\author{Fei Yan \\
University of Surrey \\
Guildford, UK \\
{\tt\small f.yan@surrey.ac.uk}
\and
Krystian Mikolajczyk \\
Imperial College London \\
London, UK \\
{\tt\small k.mikolajczyk@imperial.ac.uk}
\and
Josef Kittler \\
University of Surrey \\
Guildford, UK \\
{\tt\small j.kittler@surrey.ac.uk}}

\maketitle

\begin{abstract}

In this paper we propose a new approach to person re-identification using images and natural language descriptions. We propose a joint vision and language model based on CCA and CNN architectures to match across the two modalities as well as to enrich visual examples for which there are no language descriptions. We also introduce new annotations in the form of natural language descriptions for two standard Re-ID benchmarks, namely CUHK03 and VIPeR. We perform experiments on these two datasets with techniques based on CNN,  hand-crafted features as well as LSTM for analysing visual and natural description data. We investigate and demonstrate the advantages of using natural language descriptions compared to attributes as well as CNN compared to LSTM in the context of Re-ID. We show that the joint use of language and vision can significantly improve the state-of-the-art performance on standard Re-ID benchmarks.

\end{abstract}

\section{Introduction} 
\label{sec:intro}

%%%%%%%%%%%%%%%%%%%%%%%%%%%%%%%%%%%%%%%%%%%%

 Person Re-IDentification (Re-ID) task of matching humans across different camera views has its main application in surveillance and security. 
%With the rapidly increasing number of surveillance cameras and increasing volume of data generated by these cameras, person Re-ID has attracted a lot of research interests in the computer vision (CV) community in the last few years. 
Due to the increasing interest in largely unexplored but rapidly growing volume of visual surveillance data, much progress has recently been made with several benchmark data released. The benchmarks typically focus on classifying pairs of cropped pedestrian images as positive if they show the same person. With that clearly defined scenario, novel methods were developed in hand-crafted visual features~\cite{zhao13iccv,liao15cvpr,matsukawa16cvpr}, distance metric learning~\cite{kostinger12cvpr,mignon12cvpr,liao15cvpr}, and convolutional neural networks (CNNs)~\cite{li14cvpr,ahmed15cvpr,varior16eccv_gated}. 

\begin{figure}
  \centering
  \hspace{-2mm}
  \begin{tabular}{ c m{62mm} } \vspace{+1mm}
    \begin{minipage}{.07 \textwidth}
      \includegraphics[height=36mm]{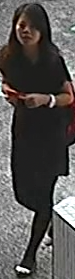}
    \end{minipage}
    & \vspace{-1mm}
 \small   An Asian girl with long, black and brown, wavy hair that reaches her shoulders. She is wearing a long dark top which is half-sleeved, with a pair of dark coloured leggings or tights, and black ballerina pumps with white designs. She carries a red or brown coloured backpack on her back. She is holding something red coloured in both of her hands, perhaps her cell phone or purse. She wears a white or sliver coloured watch or bracelet on her left wrist. 
  \end{tabular}
  \vspace{-2mm}
  \caption{An example of natural language descriptions.}
  \label{fig:example_desc_overall}
\end{figure}

However, this scenario assumes that a visual example of an ID is already available, which can then serve to query a large volume of visual data. In many practical cases such an example is not available or tedious manual search has to be done to identify one instance that can then be used as a query. In such cases a natural language description is the only available one, often gathered from a number of witnesses with many variations and inconsistencies. Examples of natural descriptions can be often found in missing person sections in newspapers (see Figure~\ref{fig:example_desc_overall}). This is a very frequent scenario, yet it has not been considered by the Re-ID community and no datasets are available to train and test suitable methods. 

Compared to attribute based annotations~\cite{layne12bmvc,layne13book}, the advantage of natural language description is its flexibility and richness. Discriminative and unique details can be captured in natural descriptions ({\em ``black ballerina pumps with white designs''}) in contrast to a predefined set of attributes. Furthermore, natural descriptions can include indications of a degree of certitude or ambiguities which may also be analysed to preserve as much information as possible ({\em ``red or brown coloured backpack'', ``perhaps her cell phone or purse'', ``white or silver coloured watch or bracelet''}). A challenge however, is to extract all the information available in this rich but unstructured description.

Owing to recent advances in computer vision, natural language processing  and machine learning, joint modelling of vision and natural language is finding more and more applications, e.g., caption generation for image and video~\cite{donahue15cvpr}, natural language based object retrieval~\cite{hu16cvpr}, zero-shot visual learning using purely textual description~\cite{elhoseiny13iccv}, and visual question answering~\cite{agrawal15vqa}. There are therefore solid foundations to introduce vision and language based techniques into the area of person Re-ID, which so far strongly relied on 
vision based methods with  excellent results on standard benchmarks  but language has been little explored in this context and this is the main goal of this work.

In this paper, we define new tasks of person Re-ID using both visual and natural language descriptions and propose new approaches to these tasks by building on recent advances in both areas of research:

\begin{itemize}

  \item We propose models that integrate vision and language, and demonstrate that in several Re-ID scenarios, the  performance can be significantly improved by the proposed integration;
  \item  We extend standard Re-ID benchmarks, CUHK03~\cite{li14cvpr} and VIPeR~\cite{gray07viper} with natural language descriptions, which will facilitate research in person Re-ID with joint vision and language modelling;
  \item We compare natural language annotations to attribute based ones, and identify their relative advantages.
\end{itemize}

The remainder of this paper is organised as follows. In Section~\ref{sec:related} we review existing person Re-ID techniques and work on joint vision and language modelling. We then introduce the descriptions we collected that allow new tasks in Section~\ref{sec:annotation}. In Section~\ref{sec:reid_with_language} we present our models and results for CUHK03. In Section~\ref{sec:lang_and_attr} we compare natural language descriptions and attributes on the VIPeR dataset. Finally, Section~\ref{sec:conclusions} concludes the paper.

\section{Related work} 
\label{sec:related}

%%%%%%%%%%%%%%%%%%%%%%%%%%%%%%%%%%%%%%%%%%%%

{\bf \noindent Person Re-Identification }
%To the best of our knowledge, the term ``person Re-Identification'' was first introduced in~\cite{gheissari06cvpr}, although the same problem had been studied in the context of multi-camera video surveillance much earlier~\cite{wang13prl,zheng16arxiv}. The last three years have witnessed an upsurge of interest in person Re-ID, thanks to the fast growing number surveillance cameras. 
Early approaches to person Re-ID were focused on improving hand-crafted features such as SIFT, LBP and histograms in  colour spaces, followed by various distance metrics learning. In~\cite{pedagadi13cvpr}, histograms in HSV and YUV spaces are considered, while in~\cite{liao15cvpr} a novel local maximal occurrence representation (LOMO) is proposed, which improves the robustness to viewpoint and illumination changes. Other recent contributions include saliency based features~\cite{zhao13cvpr,zhao13iccv,yang14eccv} and a hierarchical Gaussian descriptor~\cite{matsukawa16cvpr}.  

In terms of distance metric learning, a method based on equivalent constraints is proposed and  successfully applied to face recognition and person Re-ID in~\cite{kostinger12cvpr}. This approach is extended in~\cite{liao15cvpr} to cross-view metric learning, where the target subspace is learned by solving a generalised eigen-decomposition problem. This cross-view quadratic discriminate analysis (XQDA) has become a popular metric learning for person Re-ID. However, metric learning methods suffer from typically small data size in this application. This challenge was addressed by exploiting positive semidefinite (PSD) constraints~\cite{liao15iccv} or null space learning~\cite{zhang16cvpr_null}. Other recent metric learning techniques for Re-ID include $\ell_1$ graph~\cite{kodirov16eccv},  spatial constraints~\cite{chen16cvpr}, sample-specific or locality based learning~\cite{zheng15cvpr,zhang16cvpr_svm,li13cvpr_adaptive}, as well as learning based on low rank embedding~\cite{su15iccv}.

Following the success of deep convolutional neural networks in image classification, one of the first attempts in Re-ID was in~\cite{li14cvpr} with architecture based on pairwise data and a match/non-match binary softmax loss. A new CNN layer introduced in~\cite{ahmed15cvpr} computes cross-input neighbourhood differences, which captures local relationships between an image pair. In~\cite{xiao16cvpr}, a CNN is trained with data from multiple domains i.e. datasets, and a domain guided dropout mechanism is employed to improve the learned representation. A CNN architecture that explicitly considers multiple body parts is developed in~\cite{cheng16cvpr} with a novel triplet based loss. To deal with large intra-class variations, a metric weight constraint is incorporated into CNN training   in ~\cite{shi16eccv}. The long short-term memory (LSTM) siamese architecture has also been experimented with in~\cite{varior16eccv_lstm} to model the relations between image regions. Finally, \cite{varior16eccv_gated} applies the idea of gating to CNNs and achieves the state-of-the-art performance on standard benchmarks.

Along with the methods for person Re-ID, several datasets were introduced to the community. These include  VIPeR~\cite{gray07viper}, CUHK03 and its predecessors by \cite{li14cvpr}, Market1501~\cite{zheng15iccv_market}, PRW~\cite{zheng16arxiv_wild}, which  mainly differ by the number of cameras, IDs and data samples. A number of benchmarks were also introduced or annotated for pedestrian attribute analysis including PRID~\cite{kostinger12cvpr}, GRID~\cite{liu12eccvw}, VIPeR~\cite{layne12bmvc}, PETA~\cite{deng14peta} and RAP~\cite{li16richly}. The number and type of attributes vary which are typically predefined and binary. None of the available datasets includes natural language descriptions.

\vspace{1.5mm}
{\bf \noindent Vision and language }
Simultaneous progress in the fields of computer vision, natural language processing and machine learning has led to impressive results in integrated modelling of vision and language. Automatic image and video captioning~\cite{donahue15cvpr,karpathy15cvpr,xu15arxiv}, sentence based image retrieval~\cite{hodosh13jair} or image generation from sentences~\cite{chang15arxiv_text23d} have shown unprecedented results, which claimed to be comparable to a three-year old child. %\footnote{\url{http://www.ted.com/talks/fei_fei_li_how_we_re_teaching_computers_to_understand_pictures}}. 
Other examples of integrated modelling of vision and language include face recognition from caption-based supervision~\cite{guillaumin12ijcv}, text-to-image coreference~\cite{kong14cvpr}, zero-shot visual learning using purely textural description~\cite{elhoseiny13iccv}, and visual question answering~\cite{agrawal15vqa}. Despite those successes as well as realistic and practical case scenarios of using natural language descriptions in person Re-ID problem,  vision and language modelling has not yet been explored in this context.

\begin{figure*}
\centering
\subfigure[]{\includegraphics[height=36mm]{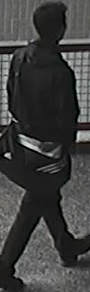}} \hspace{-1mm}
\subfigure[]{\includegraphics[height=36mm]{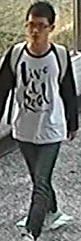}} \hspace{-1mm}
\subfigure[]{\includegraphics[height=36mm]{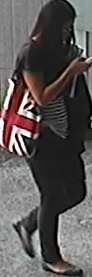}} \hspace{-1mm}
\subfigure[]{\includegraphics[height=36mm]{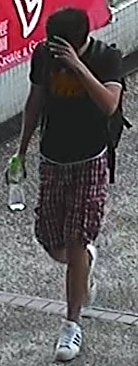}} \hspace{-1mm}
\subfigure[]{\includegraphics[height=36mm]{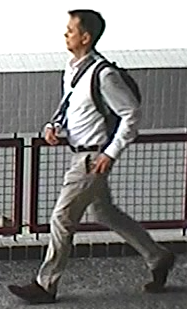}} \hspace{-1mm}
\subfigure[]{\includegraphics[height=36mm]{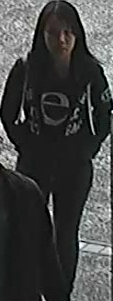}} \hspace{-1mm}
\subfigure[]{\includegraphics[height=36mm]{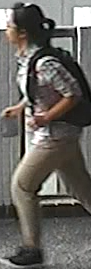}} \hspace{-1mm}
\subfigure[]{\includegraphics[height=36mm]{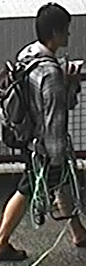}} \hspace{-1mm}
\subfigure[]{\includegraphics[height=36mm]{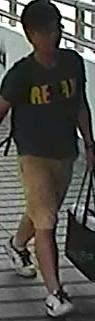}} \hspace{-1mm}
\subfigure[]{\includegraphics[height=36mm]{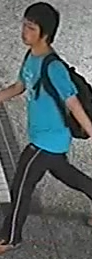}} \hspace{-1mm}
\subfigure[]{\includegraphics[height=36mm]{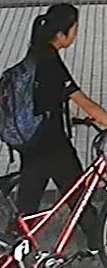}}
\vspace{-1mm}
\small
\begin{enumerate} \itemsep-0.4em
  \item[] \hspace{3mm} (a) \hspace{1mm} $\dots$ cross body bag which looks like from the brand Adidas $\dots$
  \item[] \hspace{3mm} (b) \hspace{1mm} $\dots$ that has black coloured text reading ``Live Real'' printed on it $\dots$
  \item[] \hspace{3mm} (c) \hspace{1mm} $\dots$ carrying a UK flag printed hand bag $\dots$
  \item[] \hspace{3mm} (d) \hspace{1mm} $\dots$ carrying a small transparent bottle with a green cap $\dots$
  \item[] \hspace{3mm} (e) \hspace{1mm} $\dots$ has a bandage on his right hand with a strap from his neck as he might have had an accident $\dots$
  \item[] \hspace{3mm} (f) \hspace{1mm} $\dots$ a black casual tee with a big ``e'' printed on it $\dots$
  \item[] \hspace{3mm} (g) \hspace{1mm} $\dots$ is seen rushing $\dots$
  \item[] \hspace{3mm} (h) \hspace{1mm} $\dots$ seems to be carrying an interesting metal object with green pipes dangling $\dots$
  \item[] \hspace{3mm} (i) \hspace{1mm} $\dots$ which has the text ``RELAX'' written on it in yellow $\dots$
  \item[] \hspace{3mm} (j) \hspace{1mm} $\dots$ is striding to the left $\dots$
  \item[] \hspace{3mm} (k) \hspace{1mm} $\dots$ pushes a red bicycle with both hands $\dots$
\end{enumerate}
\vspace{-3mm}
\caption{Example images from CUHK03 with discriminative sections of their natural language descriptions.}
\label{fig:example_desc_more}
\end{figure*}

\section{Natural language descriptions} 
\label{sec:annotation}

%%%%%%%%%%%%%%%%%%%%%%%%%%%%%%%%%%%%%%%%%%%%

The goal of this paper is to investigate the practical task of using natural descriptions for matching persons by joint modelling of vision and language. We therefore collect natural descriptions for two popular benchmarks,  CUHK03~\cite{li14cvpr} and  VIPeR~\cite{gray07viper}.  CUHK03 is one of the largest benchmarks for person Re-ID. It contains 13164 images of 1360 identities, captured by 6 surveillance cameras. Each identity is observed by 2 disjoint camera views. On average, there are 4.8 images per identity in each view. This dataset provides both manually labelled and automatically detected pedestrian bounding boxes. 

To collect the descriptions, we displayed a cropped person in labelled bounding box from each of the two views for each ID. We then asked the annotator to describe the person with a few sentences, that would include details to identify the person. Note that while the automatically detected bounding boxes may be less accurate than the labelled ones, there is little difference in the content of natural language descriptions. The annotation process above resulted in 2720 descriptions, one for each identity/view combination.
\begin{figure}
\centering
\subfigure[CUHK03]{\includegraphics[width=42mm]{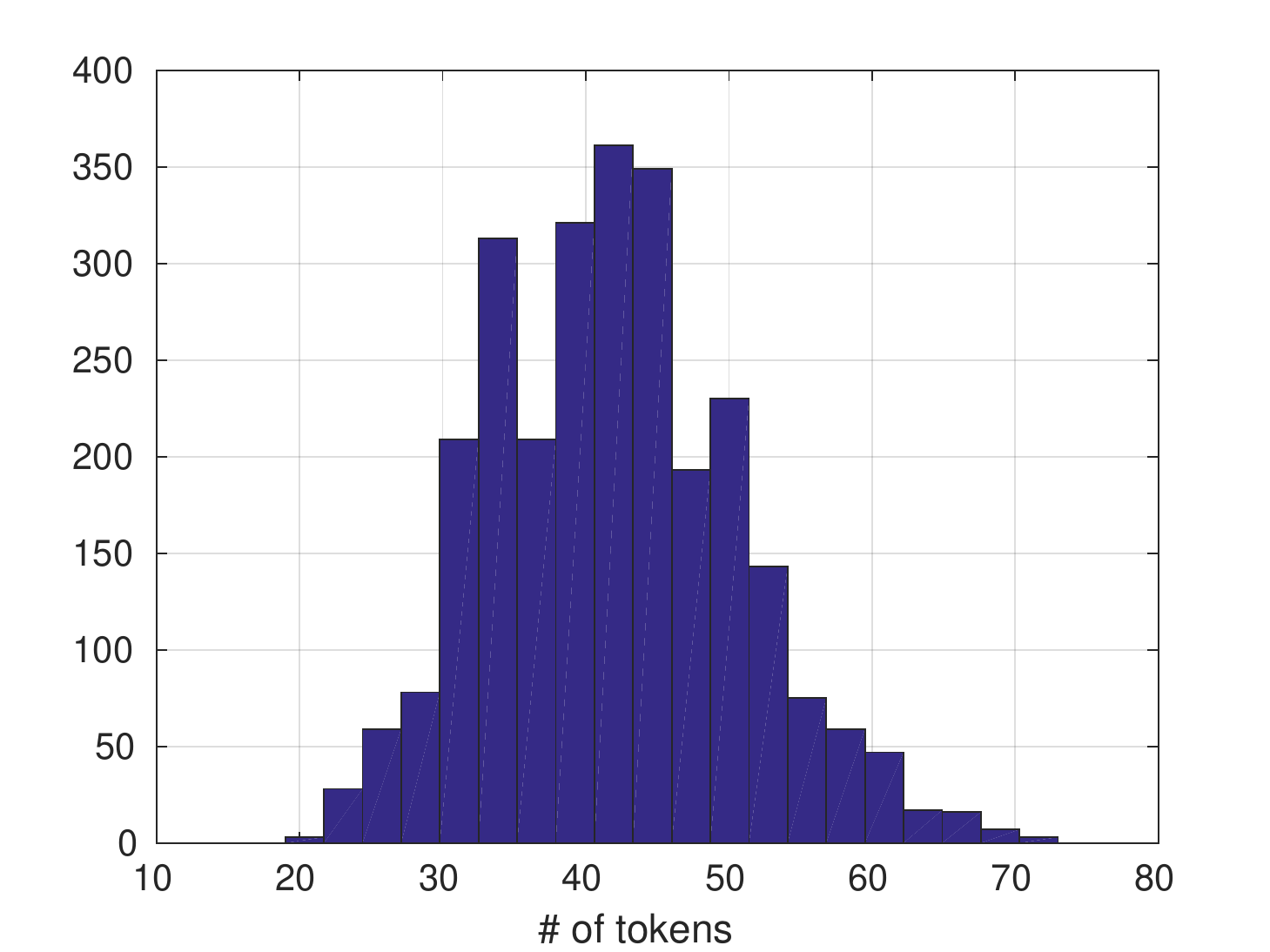}} \hspace{-3mm}
\subfigure[VIPeR]{\includegraphics[width=42mm]{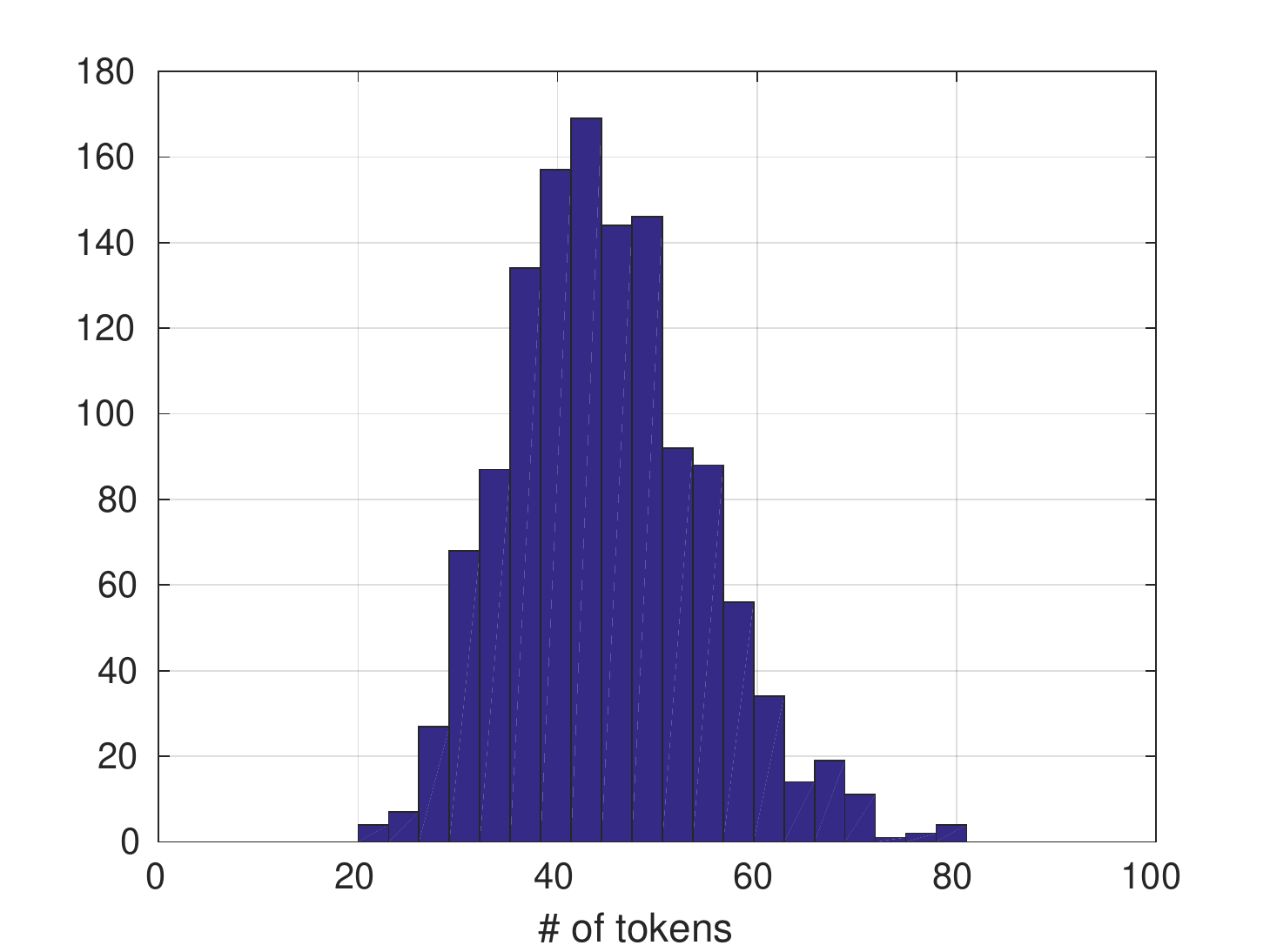}}
\vspace{-4mm}
\caption{Numbers of words in the descriptions.}
\label{fig:token_num_distr}
\end{figure}
Examples of the collected descriptions are in Figure~\ref{fig:example_desc_overall} and Figure~\ref{fig:example_desc_more}, where we show only the parts that  are discriminative but are unlikely to be included in a predefined set of attributes, thus typically available via free-style descriptions only. These include fine details e.g. (a) (c) (d) (i), 
%({\em ``from the brand Adidas'', ``text `RELAX' written on it'', ``UK flag printed hand bag'', ``transparent bottle with a green cap''}), 
unusual objects in (e) (h) (k),
%({\em ``bandage on his right hand'', ``interesting metal object with green pipes dangling'', ``a red bicycle''}), 
gait information in (g) (j), 
%({\em ``is seen rushing'', ``is striding''}), 
level of certitude ({\em ``which seems red'', ``possibly paper''}) 
or ambiguity ({\em ``white or silver'', ``watch or bracelet''}). On average, there are approximately 45 words in each description. The distribution of numbers of words is given in Figure~\ref{fig:token_num_distr}~(a).

\begin{figure*}
\centering
\subfigure[CNN] {\includegraphics[height=26mm]{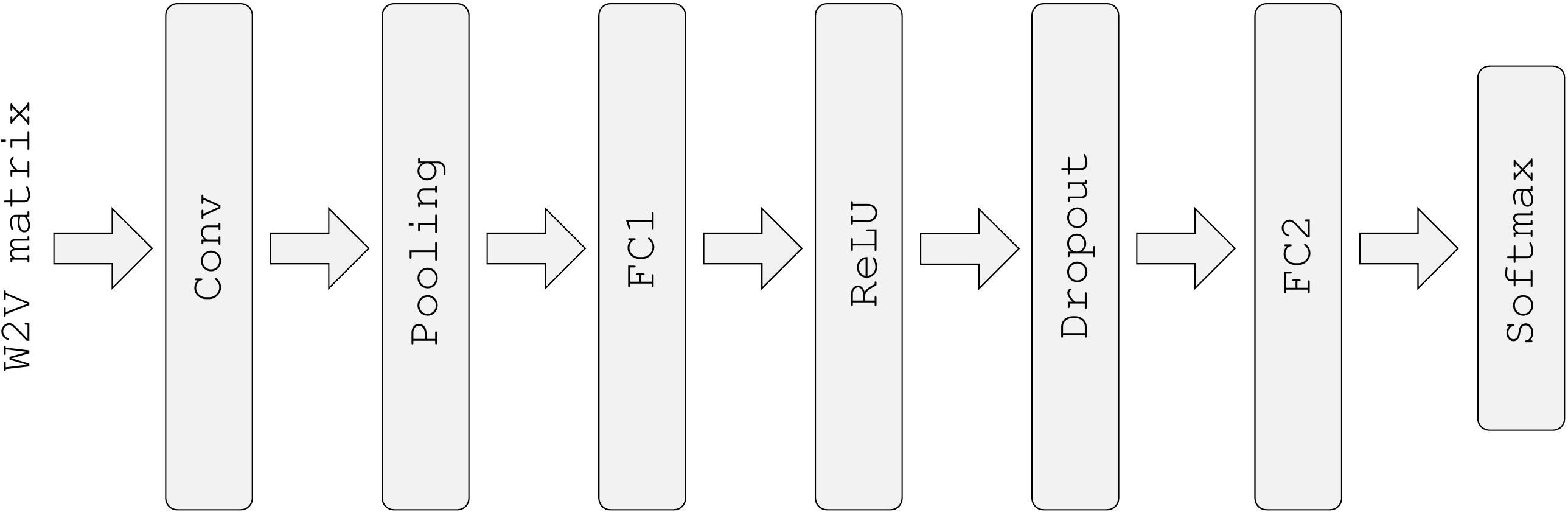}} \hspace{8mm}
\subfigure[LSTM]{\includegraphics[height=26mm]{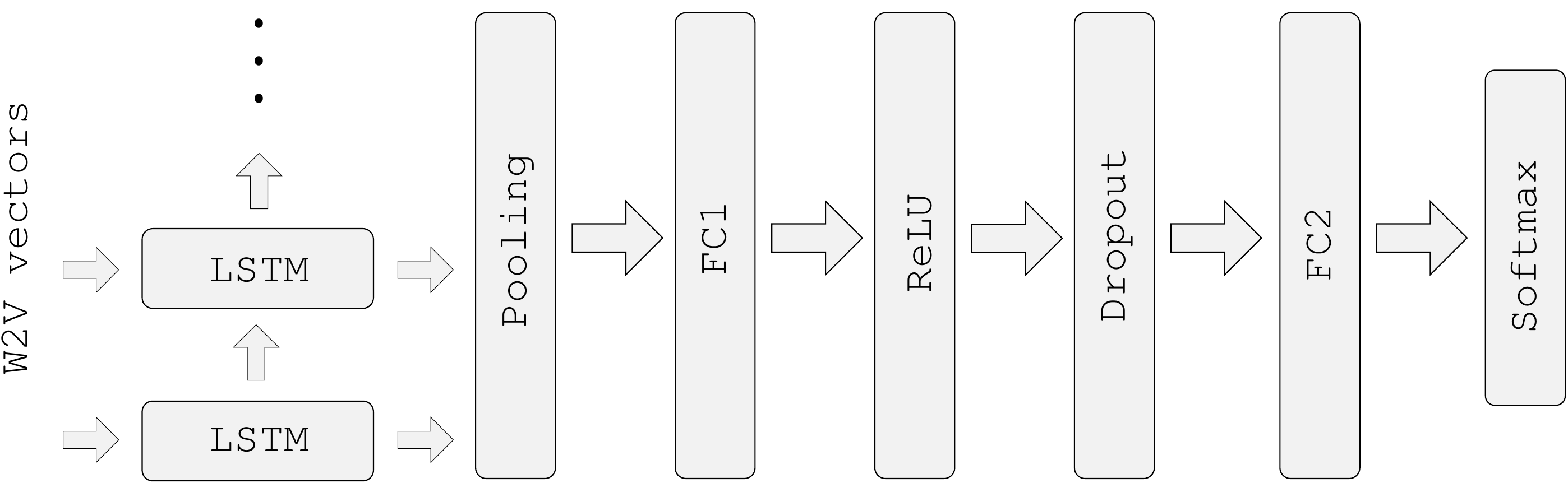}}
\vspace{-3mm}
\caption{Network architectures for language based Re-ID.}
\label{fig:text_network_arch}
\end{figure*}

The VIPeR dataset is one of the first benchmarks introduced for person Re-ID and frequently used in reported evaluations. Nevertheless, it remains one of the most challenging datasets due to the small training set size. It consists of 1264 images of 632 subjects captured with 2 cameras.   We  collected descriptions for the 1264 images in a similar way to CUHK03. The distribution of numbers of words is given in Figure~\ref{fig:token_num_distr}~(b). In total 3984 descriptions were collected for both datasets. %The annotation was done with crowd-sourcing
% through Freelancer\footnote{\url{http://www.freelancer.co.uk}}, and eight annotators from six countries were employed. 
%We did not specify the variety of the English language to use. As a result, 
British as well as American English were used by eight annotators hired via crowd-sourcing websites.% from six countries

\section{Re-Identification with vision and language} 
\label{sec:reid_with_language}

%%%%%%%%%%%%%%%%%%%%%%%%%%%%%%%%%%%%%%%%%%%%

It is evident from the images and collected descriptions (cf. Figure~\ref{fig:example_desc_more}) that they both contain rich information that is useful for person Re-ID. However, it is not straightforward to extract such information from the text due to the unstructured nature of natural language. In this section, we first present our vision only based model, then propose models for language based Re-ID, followed by the integration of vision and language. 

We also present experimental results of various settings, using the CUHK03 dataset as a testbed. We use the 1260 training identities for training and the remaining 100 for testing, and report performance averaged over the 20 predefined train/test splits. Model selection is performed separately by splitting the training set.

\subsection{Vision based Re-ID}
\label{sec:vision}
While for person Re-ID it seems natural to adopt a pairwise training scheme with a verification loss e.g. a binary softmax or a contrastive loss to drive a siamese network~\cite{li14cvpr,ahmed15cvpr,varior16eccv_gated}, in~\cite{zheng16arxiv_wild,zheng16arxiv_ide,zhong17arxiv} a classification setting is considered, and the goal is to discriminate each identity from all other identities. Compared to the verification based models, the proposed identity-discriminate embedding (IDE) has the advantage of using all available label information jointly. Once a classification network is trained, embeddings at a certain layer can be used as a representation to perform more specific tasks such as verification.

We adopt this approach by taking the pre-trained ResNet-50 model~\cite{he16cvpr_resnet}\footnote{\url{https://github.com/KaimingHe/deep-residual-networks}}, and finetuning it with the CUHK03 training data. The last fully-connected layer is modified to have 1260 outputs to match 1260 training IDs. We use the deep learning framework Caffe~\cite{jia14arxiv_caffe} and NVIDIA Titan X GPUs to train the network. We adopt a batch size of 16. For finetuning the base learning rate is set to a low value of 0.001, which drops by a factor of 0.1 every 30000 iterations. The momentum is set to 0.9 and the weight decay to 0.0005. In our experiments, we use the detected bounding boxes, rescale the images to $256 \times 256$, and augment the data with random crops of size $224 \times 224$. Once the network is finetuned, the 2048-D input to the last fully-connected layer is extracted as the IDE representation. We then employ the XQDA~\cite{liao15cvpr} to learn a discriminative metric for the final matching.

\subsection{Language based Re-ID}
\label{sec:language}
{\bf \noindent Word2Vec representation }
Each natural language description is converted from a continuous stream of text into words (or tokens) via tokenisation. %Each word is then mapped to a fixed-length representation by embedding it into a vector space.
 In contrast to Bag-of-Words that ignores the semantic correlation of words, distributed representations such as Word2vec~\cite{mikolov13nips} encode each word in a real-valued vector that preserves semantic similarity. The idea is that words that have similar contexts should be close to each other in the embedding space such that a mathematical operation on 3 vectors "king-man+woman" will result in a vector similar to "queen". Using a two-layer neural network, words are modelled based on their context, defined as a window that spans both previous and following words. We use the models\footnote{\url{https://code.google.com/archive/p/word2vec/}} pre-trained on part of Google News dataset (about 100 billion words). The model contains 300-D vectors for 3 million words and phrases.

\vspace{1.5mm}
{\bf \noindent Data augmentation }
Similarly to vision based Re-ID we adopt the identification setting of IDE for language model. There are 1260  ID classes, each with only two training instances (descriptions), one for each camera view. We consider three methods for language data augmentation. In the first one, we use synonyms to replace words in the descriptions, thus generate new variants similarly to~\cite{zhang15nips} . Based on WordNet,  the probability of choosing a synonym is larger when it is ranked  close to the most frequently seen meaning. 
The second method is simply dropping randomly words in the descriptions. Based on experiments we settled on dropping $D$ words, which is chosen uniformly from the set $\{0,1,\cdots, 10\}$. 
Finally, since the words are represented as real-valued word2vec embeddings, we also used Gaussian noise for data augmentation with a zero-mean and standard deviation of 0.05. With each of the three methods, we augment the training data by a factor of 500 thus resulting in 1000 descriptions per example.

\vspace{1.5mm}
{\bf \noindent NLP networks }
We employ two neural networks for language based Re-ID, with their architectures presented in Figure~\ref{fig:text_network_arch}. With each word encoded as a 300-D word2vec vector, we represent each description as a $300 \times 70$ matrix, where descriptions with more than 70 words are truncated, otherwise zero-padded. The CNN in Figure~\ref{fig:text_network_arch}~(a) takes the word2vec matrices as input, and convolves them with 256 kernels of size $300 \times 5$, without padding at the borders. The output is max-pooled for each channel, and is processed through a sequence of layers. The first fully-connected layer FC1 has 1024 channels, the dropout layer has a dropout ratio of 0.5, and the FC2 layer has 1260 channels, as required by the number of training IDs in the softmax loss.

We also experiment with a network based on long short-term memory (LSTM), which is illustrated in Figure~\ref{fig:text_network_arch}~(b). 
%Ignoring the bias terms, the LSTM operates as described in Eq.~\eqref{eq:lstm}, where ${\bf x}_t$, ${\bf h}_t$ and ${\bf c}_t$ are input, output and cell state vectors respectively; $A_f$, $A_i$, $A_o$, $B_f$, $B_i$ and $B_o$ are parameters matrices; and ${\bf f}_t$, ${\bf i}_t$ and ${\bf o}_t$ are the forget, input and output gates respectively, and are all learnable in the training. 
%\begin{eqnarray}
%{\bf f}_t &=& \sigma_g (A_f {\bf x}_t + B_f {\bf h}_{t-1})    \nonumber       \\
%{\bf i}_t &=& \sigma_g (A_i {\bf x}_t + B_i {\bf h}_{t-1}) \nonumber \\
%{\bf o}_t &=& \sigma_g (A_o {\bf x}_t + B_o {\bf h}_{t-1}) \nonumber \\
%{\bf c}_t &=& {\bf f}_t \circ {\bf c}_{t-1} + {\bf i}_t \circ \sigma_c (A_c {\bf x}_t + B_c {\bf h}_{t-1}) \nonumber \\
%{\bf h}_t &=& {\bf o}_t \circ \sigma_h({\bf c}_t)
%\label{eq:lstm}
%\end{eqnarray}
With the gating mechanism, LSTMs are capable of remembering long term dependencies and are particularly suitable for sequential data~\cite{hochreiter97nc}. The input to our LSTM is 300-D word2vec vectors, and the LSTM cell state vector 
%${\bf c}_t$ 
has 128 dimensions. The output of the LSTM layer is max-pooled over time, and is then processed through the same sequence of layers as in the CNN architecture. The only difference is that there are 2048 channels in FC1 rather than 1024, which were empirically chosen to give better results.

Once the networks are trained, the input to the FC2 layer is taken as features for XQDA learning and matching. For both networks the batch size is set to 100. For the CNN we use the stochastic gradient descent (SGD), and a base learning rate of 0.01, which drops by a factor of 0.1 every 50000 iterations. All other solver hyper-parameters are the same as for the vision CNN (cf. section~\ref{sec:vision}). For the LSTM, the RMSProp solver is employed, for which the base learning rate is 0.0005, the learning rate policy is ``inv'' with a gamma of 0.0001 and a power of 0.75, the RMS decay is set to 0.98, and the weight decay to 0.0005.

\vspace{1.5mm}
{\bf \noindent CNN vs. LSTM }
To gain more insight into the representation captured by the two language focused networks, we consider the case of {\em spectacles} detection as an example. {\em Spectacles} are described in our descriptions with various synonyms, to name a few, {\em glasses, sunglasses, spectacles, bespectacled, eyewear}. We expect that after training, one of the 256 channels in the Conv layer of the CNN becomes a detector for the concept {\em spectacles}, regardless of the actual synonym used. Similarly, one of the 128 channels in the LSTM layer becomes such a detector.

To verify this hypothesis, we investigate and compare the outputs of the Conv layer and the LSTM layer. Consider only the descriptions for the first view, and consider Conv layer in the CNN, 
%where $300 \times 70$ input convolved with 256 kernels of size $300 \times 5$ gives 
which outputs $F$ as a $256 \times 66$ matrix, as 2 are lost at the borders (see NLP Networks above). For each of the $256$ channels $c$ applied to a description of ID $i$, we find the index of their maximal response $u_i^c=\argmax_u F(i,c,u)+2$, where $u_i^c \in \{ 1,\cdots,70 \}$. This indicates the word within description $i$ for which channel $c$ works as a good detector. 

Next we identify the channels that are good detectors for our synonyms to {\em spectacles}.
Let $v_i$ be the ground truth index of any of the occurring synonyms in description $i$, 
 %(assuming there is at most one in each description)
where $v_i \in \{ 1,\cdots,70 \}$. We calculate the error for each channel over descriptions for all training IDs that have any of the synonyms, and identify the channel with the smallest error i.e. $\hat{c}=\argmin_c \sum_i \abs{u_i^c - v_i}$ as a detector of the concept {\em spectacles}.
It is worth noting that the above selection process indicated one channel with zero error and another one with errors for only 2 IDs. 
This suggests that there is a redundancy in the 256 channels as some of them become detectors of the same concept. All other channels have very large errors, indicating that they are detectors of other attributes. 

\begin{figure}
\centering
\subfigure[Ground truth]{\includegraphics[width=28mm]{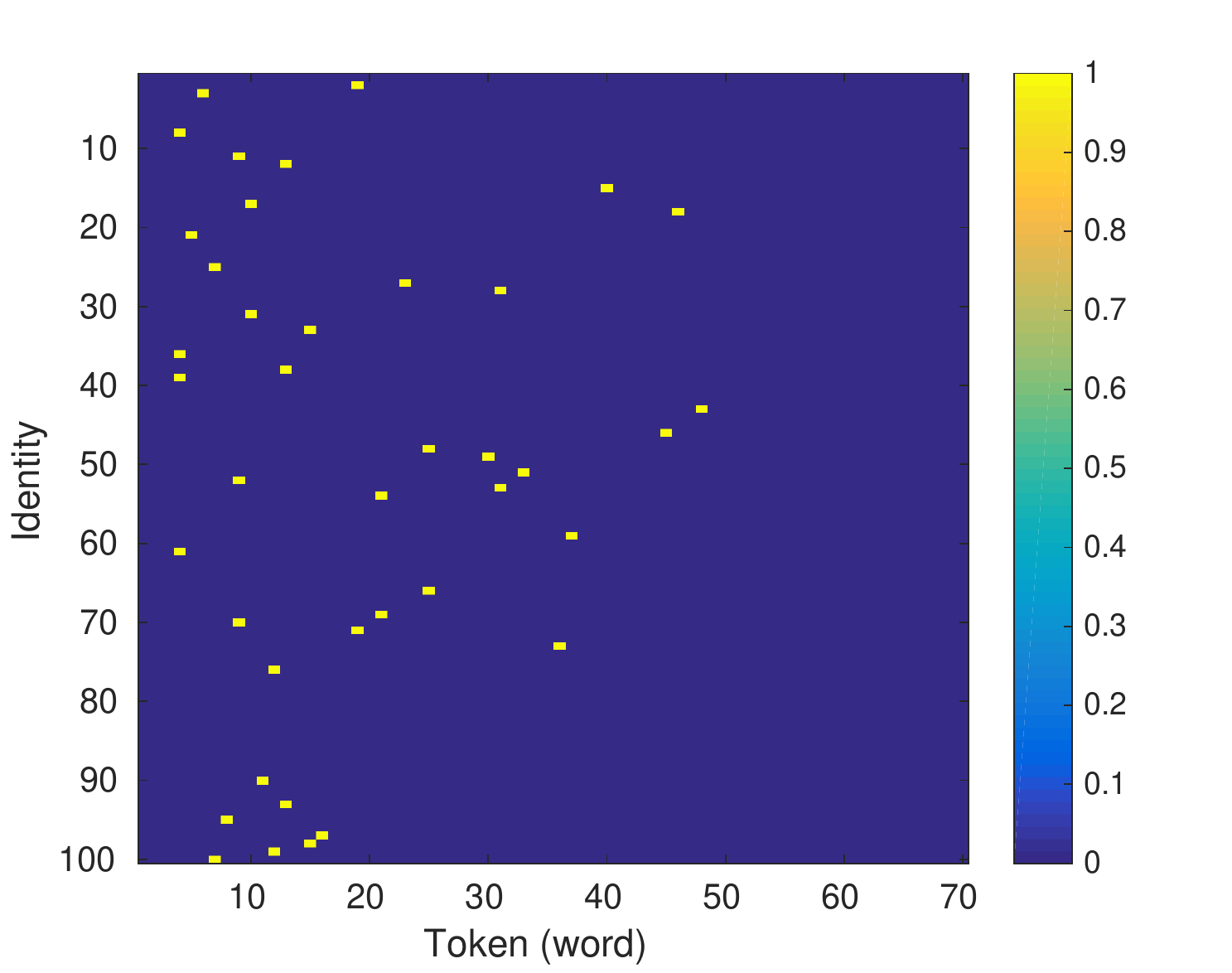}} \hspace{-3mm}
\subfigure[CNN]         {\includegraphics[width=28mm]{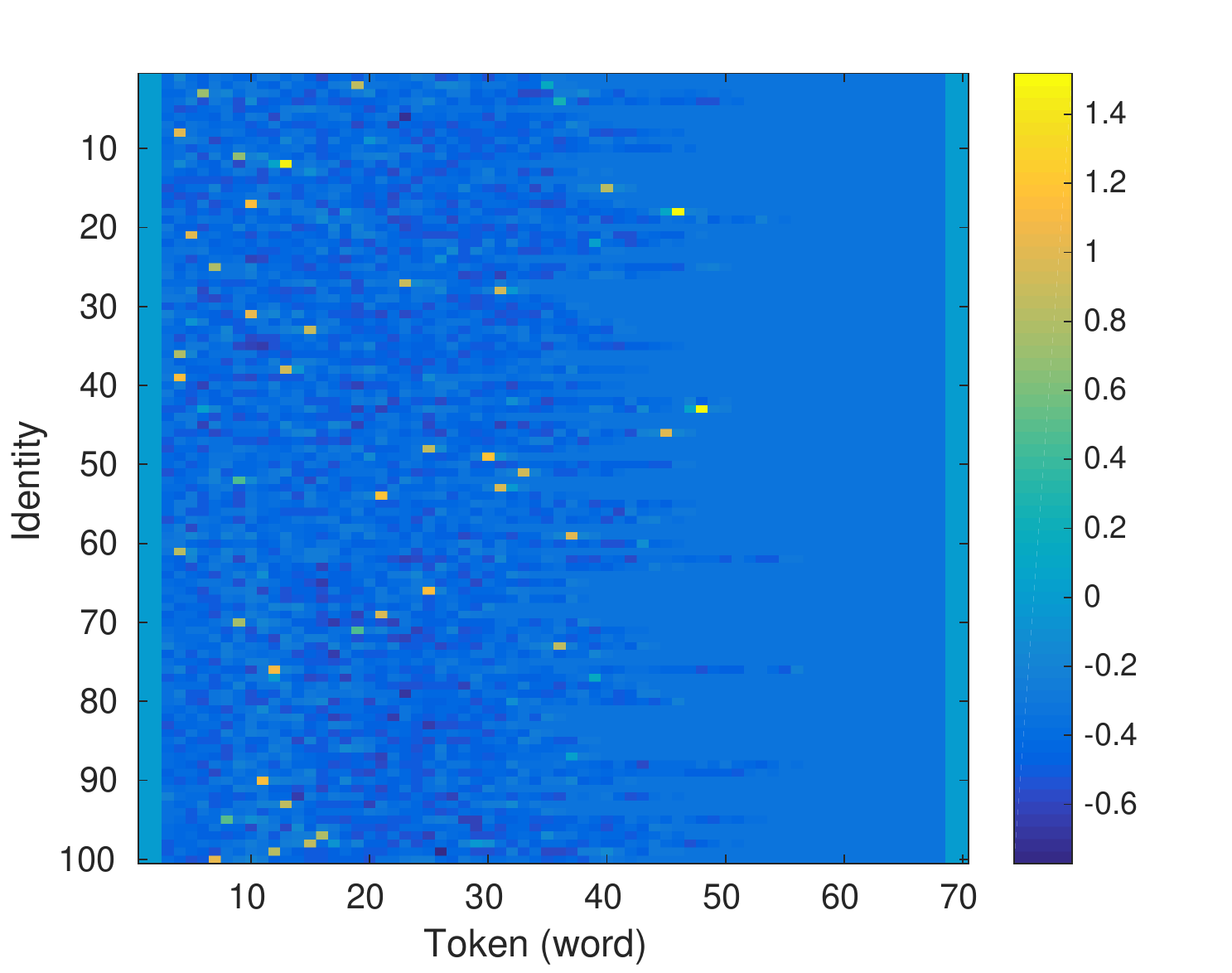}} \hspace{-3mm}
\subfigure[LSTM]        {\includegraphics[width=28mm]{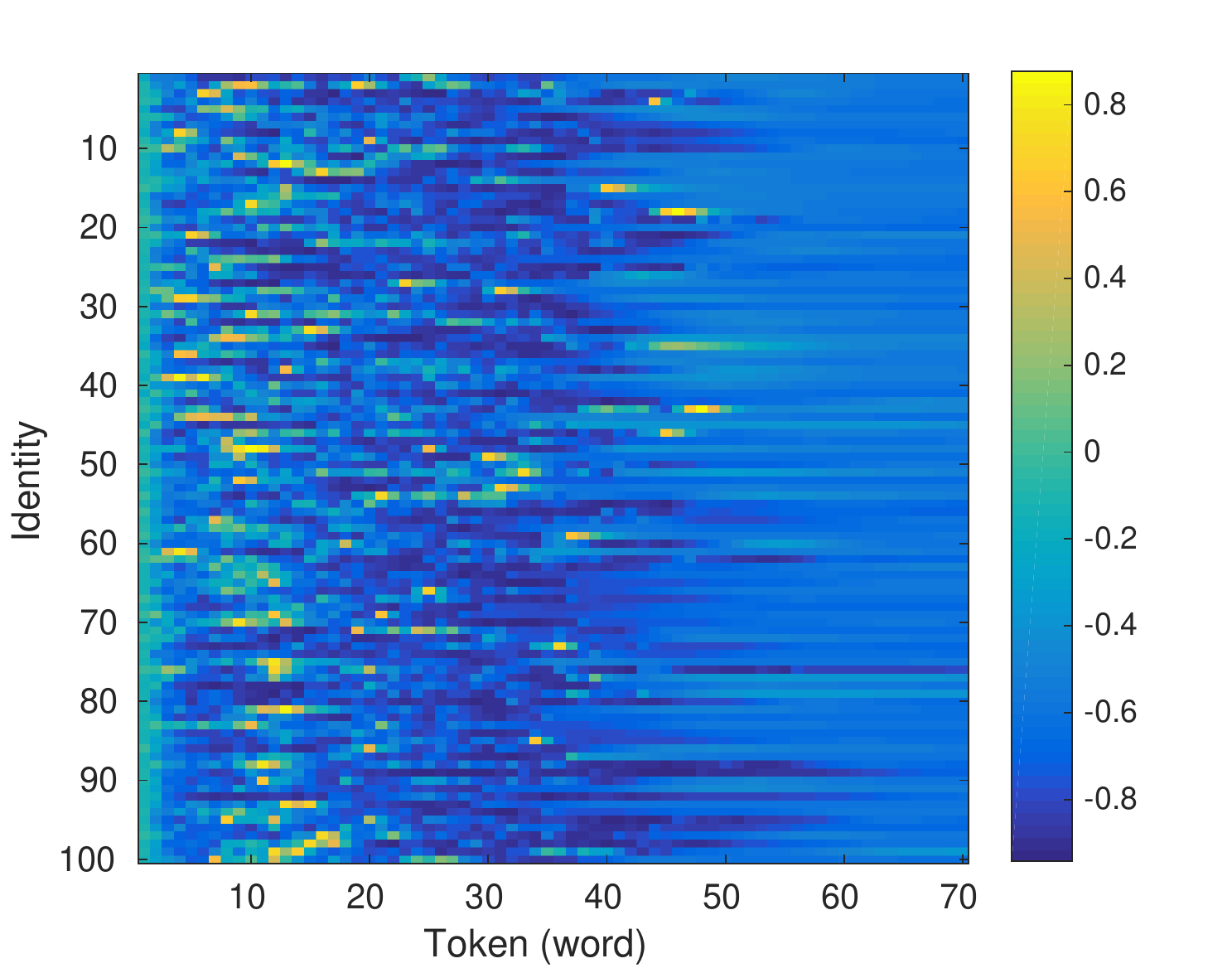}}
\vspace{-3mm}
\caption{Performance of the ``spectacles detector'' channel on the test set. This channel is identified on the training set.}
\label{fig:spectacles_detector}
\end{figure}

\begin{table}
\renewcommand{\arraystretch}{1.3}
\centering
\vspace{0mm}
\begin{small}
\begin{tabular}{ccccc}
                       & data augmentation & R1 & R5 & R10   \\
\hline
 \multirow{3}{*}{CNN}  & Synonym  &     40.0  & \bf{70.2} &     81.7  \\
\noalign{\vskip -1mm}
                       & Dropping & \bf{41.1} &     69.8  & \bf{82.5} \\
\noalign{\vskip -1mm}
                       & Gaussian &     39.4  &     68.7  &     80.7  \\ 
\hline
 \multirow{3}{*}{LSTM} & Synonym  &     38.9  &     69.6  &     79.8  \\
\noalign{\vskip -1mm}
                       & Dropping &     39.2  &     68.5  &     81.0  \\
\noalign{\vskip -1mm}
                       & Gaussian &     38.2  &     67.1  &     79.2  \\
\end{tabular}
\end{small}
\vspace{-1mm}
\caption{Rank performance on CUHK03 for  language only.}
\label{table:cuhk03_result_lang_ony}
\end{table}

The performance of the identified detector channel $\hat{c}$ on the test set is presented in Figure~\ref{fig:spectacles_detector}. Figure~\ref{fig:spectacles_detector}~(a) shows the ground truth indices of the concept {\em spectacles} with any of its synonyms used in the 100 test descriptions, and Figure~\ref{fig:spectacles_detector}~(b) and (c) plot the responses of the detector channel $\hat{c}$ in the CNN and the LSTM, respectively. Clearly the identified detector channel in CNN reliably detects attribute {\em spectacles} regardless of the actual locations of its synonym. On the other hand, the LSTM detector is considerably less accurate, with blurred detections. This indicates that this particular task is only ``weakly sequential'' and local correlation is sufficient for it, while a long term memory is harmful.
It may be possible to find optimal parameter settings for which LSTM gives better results. However, we made the same effort for both architectures and experimented with various LSTM hyper-parameters but have not been able to match the performance of CNN.

The advantage of CNN over LSTM is confirmed by the Re-ID performance in Table~\ref{table:cuhk03_result_lang_ony}, by  using features before the FC2 layer (1024-D and 2048-D for CNN and LSTM respectively).
%, and use XQDA for metric learning. this is not essential here and may be confusing
In Table~\ref{table:cuhk03_result_lang_ony}, CNN has a small but consistent edge over the LSTM, and the simple random dropping strategy works well compared to replacements by synonyms.
%{\bf needs a bit more discussion, XQDA}

\subsection{Re-ID with both vision and language}

We now investigate the integration of vision and language. For the training stage, we assume both modalities are available and for testing we consider three scenarios: 1) the gallery has only the vision modality while the query has only the language modality; 2) the gallery has only vision while the query has both vision and language; and finally, 3) both the gallery and the query have vision and language. We believe these cover the scenarios encountered in practical person Re-ID applications, and symmetric scenarios can be derived in a straightforward way.

As argued in the introduction,  vision only gallery and language only query is a frequent surveillance or missing person scenario before any image example can be found in CCTV to further query the data. Once an image is found it can be used as a typical query expansion that has been demonstrated to significantly improve retrieval results. A scenario where both language and vision are used as a query and in the gallery is also realistic for querying, for example, datasets where such descriptions are gradually created at the time of image collection e.g. dataset of missing persons or wanted offenders captured in CCTV footage. Note that many existing Police databases do use images and attributes for searching. Furthermore we may expect that in near future caption generation systems will be able to generate such language descriptions for existing datasets and make this scenario even more useful.   

\vspace{1.5mm}
{\bf \noindent Cross Modality Embedding } The first two cases are asymmetric in the sense that different modalities are available for the query  and the gallery. We employ the canonical correlation analysis (CCA)~\cite{hotelling36cca,hardoon04neuralcomputation} to bridge the modalities and ``simulate'' the missing ones. Given two sets of $m$ random vectors $X = ({\bf x}_1,\cdots,{\bf x}_m) \in \mathbb{R}^{d_x \times m}$ and $Y = ({\bf y}_1,\cdots,{\bf y}_m) \in \mathbb{R}^{d_y \times m}$, let their covariances be $\Sigma_{xx}$ and $\Sigma_{yy}$ respectively, and let the cross covariance be $\Sigma_{xy}$. CCA seeks pairs of linear projections that maximise the correlation of the two views:
\begin{eqnarray}
({\bf w}_x^*, {\bf w}_y^*) &=& \argmax_{{\bf w}_x,{\bf w}_y} \hspace{1mm} \textrm{corr}({\bf w}_x^T X, {\bf w}_y^T Y) \nonumber \\
&=& \argmax_{{\bf w}_x,{\bf w}_y} \frac {{\bf w}_x^T \Sigma_{xy} {\bf w}_y} {\sqrt{{\bf w}_x^T \Sigma_{xx} {\bf w}_x {\bf w}_y^T \Sigma_{yy} {\bf w}_y}}
\label{eq:cca_single}
\end{eqnarray}
Since the objective is invariant to scaling of ${\bf w}_x$ and ${\bf w}_y$, the projections are constrained to have unit variance:
\begin{equation}
({\bf w}_x^*, {\bf w}_y^*) = \argmax_{{\bf w}^T_x\Sigma_{xx}{\bf w}_x={\bf w}^T_y\Sigma_{yy}{\bf w}_y=1} {\bf w}^T_x\Sigma_{xy}{\bf w}_y
\label{eq:cca_single_unit_variance}
\end{equation}
Assembling the top projection vectors into the columns of projection matrices $W_x$ and $W_y$, the CCA objective can be written as:
\begin{align}
\label{eq:cca_all}
 & \hspace{5mm} (W_x^*, W_y^*) = \max_{W_x,W_y} \hspace{1mm} \textrm{tr}(W_x^T \Sigma_{xy} W_y) & \\
 & \textrm{subject to}: W_x^T \Sigma_{xx} W_x = W_y^T \Sigma_{yy} W_y = I & \nonumber
\end{align}
and Eq.~\eqref{eq:cca_all} can be solved efficiently as a generalised eigen-decomposition problem. 

In the case of person Re-ID, let $\bf x$ be the 2048-D features extracted from the vision network, as described earlier, and similarly let $\bf y$ be the 1024-D features from the text network, assuming that we use the CNN for text Re-ID. For the first scenario i.e. vision for gallery and language for query, the final gallery and query features used for metric learning and matching are:
\begin{equation}
{\bf g} = W_x^*{\bf x}, \hspace{3mm}\textrm{and}\hspace{3mm}  {\bf q} = W_y^*{\bf y}
\label{eq:feat_VxL}
\end{equation}
respectively; while for the second scenario i.e. vision for gallery and both modalities for query, the features are:
\begin{equation}
{\bf g} = {\bf x} ^\frown W_x^*{\bf x}, \hspace{3mm}\textrm{and}\hspace{3mm}  {\bf q} = {\bf x} ^\frown W_y^*{\bf y}
\label{eq:feat_VxVL}
\end{equation}
respectively, where $^\frown$ denotes concatenation of vectors. Finally, when both vision and language are available for both gallery and query, the features are simply ${\bf x} ^\frown {\bf y}$. For all three scenarios, we then learn an XQDA metric with the gallery and query features for matching.

\begin{table}
\renewcommand{\arraystretch}{1.3}
\centering
\vspace{0mm}
\begin{small}
\begin{tabular}{ccccc}
 &  & R1 & R5 & R10   \\
\hline
 \multicolumn{2}{c}{Deep ReID~\cite{li14cvpr}}              &  19.9  &     49.3  &     64.7  \\
\noalign{\vskip -1mm}
 \multicolumn{2}{c}{Unsupervised $\ell_1$ graph~\cite{kodirov16eccv}}    &  39.0    & -  & - \\
\noalign{\vskip -1mm}
 \multicolumn{2}{c}{Improved DL~\cite{ahmed15cvpr}}         &  45.0  &     75.3  &     55.0  \\
\noalign{\vskip -1mm}
 \multicolumn{2}{c}{LOMO+XQDA~\cite{liao15cvpr}}    &   46.3     &    78.9       &    88.6       \\
\noalign{\vskip -1mm}
 \multicolumn{2}{c}{Temporal adaption~\cite{martinel16eccv}}    &  48.1      &  -    & -    \\
\noalign{\vskip -1mm}
 \multicolumn{2}{c}{Sample-specific SVM~\cite{zhang16cvpr_svm}}  &  51.2  &   80.8     &  89.6   \\
\noalign{\vskip -1mm}
 \multicolumn{2}{c}{LOMO+MLAPG~\cite{liao15iccv}}   &  51.2      &   83.6        &   92.1        \\
\noalign{\vskip -1mm}
 \multicolumn{2}{c}{Deep metric embedding~\cite{shi16eccv}}  &  52.1  &  84.0   &  92.0  \\
\noalign{\vskip -1mm}
 \multicolumn{2}{c}{Single-image cross-image learning~\cite{wang16cvpr}}       &    52.2    &   84.3        &   92.3        \\
\noalign{\vskip -1mm}
 \multicolumn{2}{c}{Null space~\cite{zhang16cvpr_null}}     &  54.7  &     84.8  &     94.8  \\
\noalign{\vskip -1mm}
 \multicolumn{2}{c}{Human-in-the-loop~\cite{wang16eccv}}    &   56.1   & -     &    -  \\
\noalign{\vskip -1mm}
 \multicolumn{2}{c}{Siamese LSTM~\cite{varior16eccv_lstm}}  &  57.3  &     80.1  &     88.3  \\
\noalign{\vskip -1mm}
 \multicolumn{2}{c}{Gaussian descriptor~\cite{matsukawa16cvpr}}  &   65.5  &  88.4   &   93.7   \\
\noalign{\vskip -1mm}
 \multicolumn{2}{c}{Gated CNN~\cite{varior16eccv_gated}} &     68.1  &     88.1  &     94.6  \\
\hline
 \multirow{5}{*}{Ours}  & V x V                          &     70.3  &     93.2  &     96.6  \\
\noalign{\vskip -1mm}
                        & L x L                          &     41.1  &     69.8  &     82.5  \\
\noalign{\vskip -1mm}
                        & \hspace{-0.5mm}V x L           &     17.7  &     48.5  &     66.0  \\
\noalign{\vskip -1mm}
                        & \hspace{2mm}V x VL             &     73.5  &     94.5  &     97.7  \\
\noalign{\vskip -1mm}
                        & VL x VL                        & \bf{81.8} & \bf{98.1} & \bf{99.3} \\
\end{tabular}
\end{small}
\vspace{-1mm}
\caption{Rank at K on CUHK03, with vision and language.}
\label{table:cuhk03_results_vision_and_lang}
\end{table}

\vspace{1.5mm}
{\bf \noindent Experimental results } Evaluation results for various settings are summarised in Table~\ref{table:cuhk03_results_vision_and_lang}, where state-of-the-art results are also presented. Since the introduction of the CUHK03 dataset two years ago, the rank-1 (R1) accuracy has increased from 19.9 to 68.1. Our results are reported for various settings. For instance, ``V x V'' denotes vision for gallery and vision for query, while ``V x VL'' denotes vision for gallery and both vision and language for query.

With vision only, we achieve an R1 accuracy of 70.3, which outperforms the prior art by gated CNN (68.1). With language only, we report the best performance from Table~\ref{table:cuhk03_result_lang_ony}, i.e. with CNN and random dropping words for data augmentation. The performance is much worse than the vision only setting (41.1 vs. 70.3) as images contain more details including those that are difficult to describe with natural language. They also exhibit large illumination variations across camera views such that, the same object or clothes may appear in very different colours. This results in discrepancies in the descriptions for the different views, often more extreme than for images due to the discrete nature of language.
As expected, the ``V x L'' setting is even more challenging, with an R1 of only 17.7. However, the results for ``V x VL'' and ``VL x VL'' settings confirm that the language provides complementary information for Re-ID, with R1 scores of 73.5 and 81.8 respectively.

\section{Natural language and attributes} 
\label{sec:lang_and_attr}

%%%%%%%%%%%%%%%%%%%%%%%%%%%%%%%%%%%%%%%%%%%%

Attribute based Re-ID is closely related to language as attributes can be considered as manually extracted features from unstructured language description. We therefore compare Re-ID performance with automatic processing of language to the attributes based one on VIPeR~\cite{gray07viper} dataset. This data has been manually annotated with 15 binary attributes in~\cite{layne12bmvc}, which include: {\em shorts, skirt, sandals, backpack, jeans, logo, v-neck, open-outerwear, stripes, sunglasses, headphones, long-hair, short-hair, gender, carrying-object}, 12 of which are appearance-based, and 3 are soft-biometrics. Following the protocol we use 316 identities for training and the remaining 316 for testing, and report averaged results over 10 random splits. 

\begin{figure}
  \centering
  \hspace{-2mm}
  \begin{tabular}{ c m{67mm} }
    \begin{minipage}{.07 \textwidth}
      \includegraphics[height=34mm]{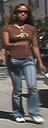}
    \end{minipage}
    & \vspace{2mm}
     A front profile of a young, slim and average height, black female with long brown hair. She wears sunglasses and possibly earrings and necklace. She wears a brown t-shirt with a golden colored print on its chest, blue jeans and white sports shoes. \vspace{10mm} \\
    \begin{minipage}{.07 \textwidth}
      \includegraphics[height=34mm]{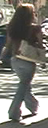}
    \end{minipage}
    & \vspace{0mm}
     A short and slim young woman carrying a tortilla coloured rectangular shoulder bag with caramel straps, on her right side. She has a light complexion and long, straight auburn hair worn loose. She wears a dark brown short sleeved top along with bell bottomed ice blue jeans and her shoes can't be seen but she might be wearing light colored flat shoes.
  \end{tabular}
  \vspace{0mm}
  \caption{Natural language descriptions and attributes for a pair of example images in VIPeR. Top: view 1; bottom: view 2. The positive attributes for this identity are: {\em jeans, sunglasses, longhair.}}
  \label{fig:example_desc_viper}
\end{figure}

In Figure~\ref{fig:example_desc_viper} we show natural language descriptions and attributes for an example pair of images in VIPeR. 
An intrinsic advantage of free-style descriptions is their level of details that cannot be predicted in advance in form of a fixed list of attributes for whole dataset, in particular if that data is growing in time. Moreover, note that in VIPeR the attributes are annotated on a per-identity basis. For instance,  in Fig.~\ref{fig:example_desc_viper}, the sunglasses are visible only in the first view, but are annotated as positive for the identity in both views. In contrast, our natural language descriptions were written independently for the two views by different annotators. 
% at least some of them where, so I'd leave this, we can add more annotations later
As a result, discrepancies exist in the descriptions due to change of pose, illumination, occlusion, and annotator. For instance, the same person is described as ``{\em black female with long brown hair}'' for one view, and as someone with ``{\em light complexion and long, straight auburn hair}'' for another view. Such descriptions do not require expertise from the annotators and can be done with less effort compared to a long list of attributes. They reflect a more realistic scenarios,  and pose interesting challenges for learning.

For vision only Re-ID we considered finetuning the ResNet-50 model. However, due to the small size of the training set in VIPeR, the achieved R1 accuracy  was below 35. Moreover, the performance of Gated CNN~\cite{varior16eccv_gated} is 37.8 in contrast to its state-of-the-art results on CUHK03. We therefore experimented with the hierarchical Gaussian descriptor (HGD)~\cite{matsukawa16cvpr}\footnote{\url{http://www.i.kyushu-u.ac.jp/~matsukawa/ReID.html}}, which was the best available feature for VIPeR to our knowledge. There are 4 versions of HGD defined in different colour spaces and each is 7567-D. We chose the RGB version which performed the best in our experiments. For language we train the same CNN as in CUHK03 (cf. section~\ref{sec:language}) with VIPeR descriptions and extract the 1024-D features.

\begin{table}
\renewcommand{\arraystretch}{1.3}
\centering
\vspace{0mm}
\begin{small}
\begin{tabular}{ccccc}
 &  & R1 & R5 & R10   \\
\hline
 \multicolumn{2}{c}{Local FDA~\cite{pedagadi13cvpr}}    &  24.1    & 51.2  & 67.1 \\
\noalign{\vskip -1mm}
 \multicolumn{2}{c}{Unsupervised salience~\cite{zhao13cvpr}}    &  26.9    & 47.5  & 62.3 \\
\noalign{\vskip -1mm}
 \multicolumn{2}{c}{Mid-level filters~\cite{zhao14cvpr}}    &  29.1    & 52.3  & 65.9 \\
\noalign{\vskip -1mm}
 \multicolumn{2}{c}{Locally-adaptive decision~\cite{li13cvpr_adaptive}}  &  29.4  & 63.3  & 76.3 \\
\noalign{\vskip -1mm}
 \multicolumn{2}{c}{Query-adaptive fusion~\cite{zheng15cvpr}}    &  30.2    & 51.6  & 62.4 \\
\noalign{\vskip -1mm}
 \multicolumn{2}{c}{Salience matching~\cite{zhao13iccv}}    &  30.2    & 52.3  & 65.5 \\
\noalign{\vskip -1mm}
 \multicolumn{2}{c}{Improved DL~\cite{ahmed15cvpr}}         &  34.8  &     63.7  &     75.8  \\
\noalign{\vskip -1mm}
 \multicolumn{2}{c}{Correspondence structure learning~\cite{shen15iccv}}    &  34.8  &  68.7 & 82.3    \\
\noalign{\vskip -1mm}
 \multicolumn{2}{c}{Single-image cross-image learning~\cite{wang16cvpr}}    &  35.8  &  67.4   &   83.5        \\
\noalign{\vskip -1mm}
 \multicolumn{2}{c}{Gated CNN~\cite{varior16eccv_gated}}         &  37.8  &     66.9  &  77.4  \\
\noalign{\vskip -1mm}
 \multicolumn{2}{c}{Salient colour names~\cite{yang14eccv}}         &  37.8  &     68.5  &  81.2  \\
\noalign{\vskip -1mm}
 \multicolumn{2}{c}{Domain guided dropout~\cite{xiao16cvpr}}       &    38.6    &   -    &  -        \\
\noalign{\vskip -1mm}
 \multicolumn{2}{c}{LOMO+XQDA~\cite{liao15cvpr}}    &   40.0     &    68.1       &    80.5       \\
\noalign{\vskip -1mm}
 \multicolumn{2}{c}{LOMO+MLAPG~\cite{liao15iccv}}   &  40.7      &   69.9        &   82.3        \\
\noalign{\vskip -1mm}
 \multicolumn{2}{c}{Semantic transfer~\cite{shi15cvpr}}    &   41.6   & 71.9     &   86.2  \\
\noalign{\vskip -1mm}
 \multicolumn{2}{c}{Low rank embedding~\cite{su15iccv}}    &   42.3   & 72.2     &   81.6  \\
\noalign{\vskip -1mm}
 \multicolumn{2}{c}{Siamese LSTM~\cite{varior16eccv_lstm}}  &  42.4  &    68.7  &    79.4  \\
\noalign{\vskip -1mm}
 \multicolumn{2}{c}{Sample-specific SVM~\cite{zhang16cvpr_svm}}  &  42.7  &   -     &  84.3   \\
\noalign{\vskip -1mm}
 \multicolumn{2}{c}{Parts-based CNN~\cite{cheng16cvpr}}  &   47.8  &  74.7   &   84.8   \\
\noalign{\vskip -1mm}
 \multicolumn{2}{c}{Gaussian descriptor~\cite{matsukawa16cvpr}}  &   49.7  &  79.7   &   88.7   \\
\noalign{\vskip -1mm}
 \multicolumn{2}{c}{Null space~\cite{zhang16cvpr_null}}            &   51.7    &   82.1  &  90.5  \\
\noalign{\vskip -1mm}
 \multicolumn{2}{c}{Spatial similarity learning~\cite{chen16cvpr}} & {\bf 53.5} &  {\bf 82.6} &  {\bf 91.5}  \\
\hline
 \multirow{4}{*}{Ours}  & V x V                          & 42.3 &  71.5 &  82.8 \\
\noalign{\vskip -1mm}
                        & L x L                          & 13.2 &  31.9 &  42.7 \\
\noalign{\vskip -1mm}
                        & VL x VL                        & 52.1 &  80.3 &  89.9 \\
\hline
  \multirow{4}{*}{Attrib., VA x VA} & N = 0              & 95.9 & 100.0 & 100.0 \\
\noalign{\vskip -1mm}
                                    & N = 1              & 61.9 &  90.8 &  95.9 \\
\noalign{\vskip -1mm}
                                    & N = 2              & 48.1 &  77.9 &  88.1 \\
\noalign{\vskip -1mm}
                                    & N = 3              & 40.4 &  72.2 &  83.4 \\
\end{tabular}
\end{small}
\vspace{-2mm}
\caption{Rank at K on VIPeR, with vision and language.}
\label{table:viper_results}
\end{table}

In Table~\ref{table:viper_results} we show our results and compare to those in the literature. The vision representation with the RGB version of HGD achieves an R1 of 42.3, which is higher than for ResNet-50 model but lower than the fused version of HGD reported in~\cite{matsukawa16cvpr} (49.7), and lower than several other methods.  This confirms that deep CNN architectures require large training data which is rarely available in practical Re-ID scenario and hand-crafted features with distance metric learning may outperform CNNs on small datasets such as VIPeR.
With the help of language, the joint Re-ID method gives an R1 of 52.1, which is still lower than the state-of-the-art method with spatial similarity learning~\cite{chen16cvpr} (53.5). However, it improves the vision baseline by approximately 10 points, which again confirms that language provides complementary information for Re-ID. In fact, by combining with language, we expect to have improvement of the same order over most existing vision based Re-ID systems.

Since the attributes are annotated for identities rather than images, using them directly as additional features to vision would lead to a perfect Re-ID performance (assuming the attributes are unique for each identity). We therefore randomly flip $N$ bits of the binary attributes, to simulate the more realistic scenario where the two views are annotated independently. It is observed from the results in the bottom section of Table~\ref{table:viper_results} that when 2 of the 15 binary bits are flipped, the performance of joint vision and attributes drops below that of vision and language.  This further supports  the advantage of natural language descriptions over attributes.

%{\bf Some final remarks that language is great for Re-ID? better in conclusions}

\section{Conclusions} 
\label{sec:conclusions}

%%%%%%%%%%%%%%%%%%%%%%%%%%%%%%%%%%%%%%%%%%%%

We have presented a new approach to person Re-ID based on joint modelling of vision and language. This reflects various practical scenarios where visual examples are not available to query a database, and language to vision search is the only alternative to a manual browsing of visual data. To that end we extended the annotations of two standard Re-ID benchmarks by collecting natural language descriptions.

We have proposed methods that integrated vision and language and improved upon the state-of-the-art results in two standard Re-ID benchmarks.  We have investigated three scenarios based on whether language and/or vision are used as a query, gallery or both and achieved significant and consistent improvements. In particular, vision complemented with language description boosts the performance by more than 10 points.

Several key observations have been made in the context of applying language and vision to Re-ID problem. CNN performed well for images and text when large training data was available otherwise hand-crafted features  gave better results. CNN has performed consistently better than LSTM for detection of attributes that can be described by synonyms. We believe that it is due to long term relations captured by LSTM that are impeding its performance in the case of Re-ID.  

Language is a less rich description than vision but provides complementary information. In particular, high level reasoning can be used and various details that are not directly observable from the given image can be inferred and included. Compared to a list of attributes, natural language descriptions have a number of advantages. They do not require expert annotators, do not assume zero annotation noise, can be easily expanded, allow for including specific and unique details, are faster to generate and lead to better performance. 

With recent progress in caption generation one can expect that such descriptions can be generated automatically and then used in combination with vision or as a proxy between camera views with extreme change of angle. Future work may include experiments with such automatically generated descriptions, using more powerful language models.

% We have also proposed and demonstrated improved performance of an approach that can make use of joint language and  vision model even if language is available only during training and not for testing.

% and trained on other visual recognition problems with larger datasets. 
 
%Future work: joint training, description generation. more powerful langauge model?

\section*{Acknowledgement}

This work has been supported by EPSRC EP/N007743/1 FACER2VM project.

\end{document}